\documentclass[sigconf]{acmart}

\renewcommand\footnotetextcopyrightpermission[1]{} 
\setcopyright{none}

\usepackage{mathtools}
\usepackage{amsmath}
\usepackage{marvosym}
\usepackage{listings}
\usepackage{dblfloatfix}
\usepackage{colortbl}
\usepackage[utf8]{inputenc}
\usepackage{float}
\usepackage{subfig}
\usepackage{multirow}
\usepackage{algorithm}
\usepackage{algorithmic}
\usepackage{pifont}
\usepackage{mathtools}
\usepackage{amsfonts}
\usepackage{xspace}
\usepackage{array}
\usepackage{enumerate}
\usepackage{graphicx}

\begin{document}

\title{Towards Enhancing Fault Tolerance in Neural Networks}

\author{Vasisht Duddu}
\affiliation{%
  \institution{University of Waterloo}
  \country{Canada}
}
\authornote{Work done at Indraprastha Institute of Information Technology (IIIT), Delhi, India}
\email{vasisht.duddu@uwaterloo.ca}

\author{D. Vijay Rao}
\affiliation{%
	\institution{Institute for Systems Studies \& Analyses}
  \country{India}
}
\email{vijayrao@issa.drdo.in}

\author{Valentina E. Balas}
\affiliation{%
	\institution{Aurel Vlaicu University of Arad}
  \country{Romania}
}
\email{valentina.balas@uav.ro}


\begin{abstract}
Deep Learning Accelerators and Neuromorphic hardware, used in many real-time safety-critical applications, are prone to faults that manifest in the form of errors in Neural Networks.
Fault Tolerance in Neural Networks is a critical attribute for applications that require reliable computation for long duration such as IoT and mobile devices.
The inherent fault tolerance of Neural Networks can be improved with regularization, however, the current techniques exhibit a trade-off between generalization and classification accuracy.
To this extent, in this work, a Neural Network is modelled as two distinct functional components: a \textit{Feature Extractor} with an unsupervised learning objective and a \textit{Fully Connected Classifier} with a supervised learning objective.
Traditional approaches to train the entire network using a single supervised learning objective are insufficient to achieve the objectives of the individual functional goals optimally.
In this work, a novel two phase framework with multi-criteria objective function combining unsupervised training of the Feature Extractor followed by supervised training of the Classifier Network is proposed.
In the Phase I, the unsupervised training of the Feature Extractor is modeled using two games solved simultaneously in the presence of Neural Networks with conflicting objectives.
The first game with a generative model, trains the Feature Extractor to generate robust features for the input image by minimizing a reconstruction loss between the input and reconstructed image.
The second game with a binary classification network, updates the Feature Extractor to smoothen the feature space and match with a prior Gaussian distribution.
In Phase II, the resultant Feature Extractor, which is strongly regularized, is combined with the Fully Connected Classifier for fine-tuning on the classification task.
The proposed two phase training algorithm is evaluated on four architectures with varying model complexity on standard image classification datasets: FashionMNIST and CIFAR10.
The proposed framework is scalable and independent of the network architecture that provides \textit{superior} tolerance to stuck at ``0" faults as compared to existing regularization functions without loss in classification accuracy.
\end{abstract}


\maketitle
\pagestyle{plain}

\section{Introduction}\label{intro}

The tremendous growth and adoption of advanced machine learning techniques, such as Deep Neural Networks (NN), is attributed to the improvements in algorithms, availability of massive data and advances in the hardware design.
However, this tremendous interest in NNs has also raised questions about the fault tolerance (FT) of AI systems which are yet to be addressed. 
Resilience to random hardware faults is an important property for AI systems deployed in mission critical applications such as autonomous vehicles and spacecrafts which demand high performance and reliability with graceful degradation.
Often, for achieving a high level of on-board automation, these systems frequently use NNs on Neuromorphic circuits and accelerators which rely on inherently unreliable semiconductor devices owing to process variations, thermal issues and leakages which introduces errors in computation \cite{doi:10.1162/0899766053723096}. Safety requirements along with functional correctness and reliability constraints need to be designed as part of the NNs for these applications. Instead of viewing the FT of Deep Learning systems from a hardware perspective, this work describes a framework based on game theory to improve the \textit{inherent} FT of the NN \textit{application} running on the hardware.

\noindent\textbf{Problems with current solutions.} Traditional reliability engineering techniques to enhance FT via explicit redundancy combined with voting strategies cannot be directly incorporated for NNs~\cite{Li:2017:UEP:3126908.3126964}.
For instance, \textit{N-Modular Redundancy} is infeasible for resource constraint applications due to computation overhead and additional synchronous communication requirement between NNs.
Alternatively, \textit{augmenting redundancy} by adding additional nodes and synapses to distribute the computational load among other nodes results in large networks further requiring post-processing techniques like pruning~\cite{doi:10.1162/0899766053723096}.
Further, identifying critical nodes in every layer and distributing the load to other nodes is not feasible for large networks with millions of parameters~\cite{8013784}.
On the other hand, FT modelled as a \textit{constraint} is computationally expensive, for instance, min-max constraint cannot be optimized using gradient descent as the function is not differentiable.
This requires approximating the objective function resulting in partial FT~\cite{105414,712162}. Hence, a viable direction to enhance the inherent FT of NNs is by improving the generalization of the model by adding regularization function to the training objective~\cite{8561200,7862272,10.1162/neco.1995.7.1.108}. However, as shown in this paper, prior regularization approaches face an unbalanced accuracy-generalization trade-off which impacts the overall FT.

\noindent\textbf{Proposed Approach.} In this work, we address the challenge of enhancing the inherent FT of NNs to random faults by improving the generalization while maintaining high accuracy. Our key idea in improving the FT \textit{through regularization} is to identify the individual NN components with different functionality and train them separately with different objectives for best performance and higher generalisation.
The NN is divided into a Feature Extractor (FE) network with an unsupervised learning objective of extracting the dominant and robust features from the input image, and a Fully Connected Classifier (FCC) Network with a supervised learning objective to predict the image label given the extracted features.
Previously proposed algorithms for FT train the entire NN using a single supervised learning objective which is insufficient to achieve the learning objectives of the individual components.
This further leads to a tradeoff between the model accuracy and generalization which affects the FT of the network.
A novel two phase multi-criteria fault tolerant training algorithm is proposed, comprising of an unsupervised objective function for the FE and a supervised objective function for training FCC.

\noindent\textbf{\underline{Phase I.}} First, we train the FE with unsupervised objective with the goal to: (a) extract robust features of a given image and (b) smoothen the intermediate feature space to match prior Gaussian distribution. Each of these goals are achieved by training the FE within a \textit{adversarial game theoretic framework} with other NNs with conflicting objectives.

\noindent\ding{202} \textbf{Game 1.} The FE is trained to extract robust features in presence of a Generative Model with conflicting objectives. The FE model maps the input images to corresponding dominant features, while the Generative model reconstructs the image \textit{given} the FE's extracted features. Both the networks are trained jointly, to minimize the reconstruction loss between the input and reconstructed images such that they are indistinguishable.

\noindent\ding{203} \textbf{Game 2.} The FE is trained to match the feature space with Gaussian distribution (smoothening intermediate representation) using a minimax objective function in the presence of a Discriminator Network. The discriminator (binary classifier) aims to differentiate between samples from Gaussian distribution and generated samples from the FE. This minimax objective acts as a FT constraint to \textit{minimize the maximum deviation} between the samples from the feature space distribution and target prior distribution.
    
\noindent\textbf{\underline{Phase II.}} The resultant FE after unsupervised training is attached to the FCC and fine-tuned to classify the features extracted by FE to a corresponding label. This is achieved by minimizing supervised loss for object classification using Gradient Descent.

\noindent\textbf{Main Contributions.}  In this work, we first show that prior state of the art regularization functions to enhance FT, i.e, \textit{Lasso (L1) and Tikhonov (L2) regularizer}, have a tradeoff between accuracy and generalisation. This tradeoff makes standard regularization functions unsuitable for incorporating FT into the network. We show that the generalized models have lower standard deviation of parameter values compared to overfitting models which is the primary reason to higher tolerance. Based on this observation, we propose generalization error (difference in training and testing accuracy) as a metric to compare the relative FT between different models by capturing overfitting (Section~\ref{reg}).

To achieve high generalisation via strongly regularising the model, a novel framework for training the NN is proposed \textit{combining unsupervised and supervised learning objectives} modelled as two adversarial games with conflicting NNs (Section~\ref{advfault}). Further, we extend stuck at ``0" faults to cover filters in convolutional NNs in addition to weights in Multilayer Perceptrons.

On comparing the proposed algorithm with other regularization approaches, the resultant NNs depict lower generalization error as well as a higher FT and test accuracy (Section~\ref{eval}). We simulate and compare the performance degradation of different models in the presence of random stuck at ``0" faults in the NN simulated with random zero values in parameters and node outputs.
The evaluation is performed on networks of varying network complexities to indicate that the \textit{training approach is scalable and independent of the model architecture}. The training is on benchmarking datasets, FashionMNIST and CIFAR10, with varying learning difficulty for the NNs. 

\section{Background and Related Work}\label{background}

\subsection{Machine Learning}

Machine Learning algorithms learn a function $f: X \longrightarrow Y$ to map the data samples in $X$ to its corresponding class in $Y$.
This is achieved by minimizing the loss $l(f(x), y)$ over predicting the output of data instance $(x, y)$, $f(x)$, compared to the ground truth label $y$. Instead of optimizing the loss over the entire data population $P(X, Y)$ (not tractable), the loss is estimated over the training dataset $D \sim P(X, Y)$. However, machine learning models have higher train accuracy compared to test accuracy (overfitting). A regularization function $R(\theta)$ is added to the objective which penalises large parameter values to fit the training dataset $D$: $\min\limits_f \, L_{D} + \lambda \, R(\theta)$.

\noindent NNs comprise of distributed network of computational units connected with each other through edges with a weight associated which indicates the importance of the input for the computation in the next layer.
Each neuron computes the weighted average of all the input synapses, $a^{l} = W^{l} \times a^{l-1} + b^{l}$ where $a^{l}$ is the activation (intermediate sum) of the $l^{th}$ layer, $W^{l}$ are weights learned for each of the synapses during training, $a^{l-1}$ are the activation values from the previous layer fed as input and bias values are given by $b^{l}$.
A non-linear activation function follows the matrix-vector computation which restricts the activation values from growing too large.

\subsection{Fault Tolerant Neural Networks}

NNs have certain degree of inherent FT to failure in nodes or loss of edges due to high degree of overparametersization~\cite{363479}. FT of NNs ensures that it continues to operate even in the presence of node and weight faults, degrading gracefully over time.
According to the definition of $\epsilon$-FT~\cite{105414}, a NN $\mathcal{N}$ performing computations $\mathcal{H}_\mathcal{N}$ is said to be fault tolerant if the computation $\mathcal{H}_{\mathcal{N}_{fault}}$ performed by a faulty NN $\mathcal{N}_{fault}$
is close to $\mathcal{H}_\mathcal{N}$, Formally,
$\mid \mid \mathcal{H}_\mathcal{N}(\mathcal{X}) -  \mathcal{H}_{\mathcal{N}_{fault}}(\mathcal{X}) \mid \mid \leq \epsilon $
for $\epsilon>0$ and input image $\mathcal{X}$ is sampled from the training dataset $D$.
For a NN to be completely fault tolerant, the value of $\epsilon=0$.
However, this strict condition of complete FT can be relaxed by designing a NN with graceful degradation where $\epsilon > 0$.
The resultant model $\mathcal{N}$ satisfying the above constraint is referred to as $\epsilon$-Fault Tolerant.

FT techniques can be broadly classified into active and passive FT (refer to~\cite{8013784} for a detailed survey).
Active approaches dynamically recognize and manage the system's redundant resources to compensate for faults as they appear, by adaptation, relearning and self repair mechanisms.
This, however, is complex with additional detection and localization logic within the system.
Generally, a better degree of FT can be achieved using passive techniques over active approaches which cannot cover all the possible cases.
In passive techniques, intrinsic redundancy and fault masking is incorporated into the models before training to ensure correct operation in the presence of faults.

FT can be incorporated as a constraint during training using Quadratic programming~\cite{105414}, Genetic Algorithms~\cite{Su2016} and minimax optimisation~\cite{712162}.
These approaches provide enhanced FT with theoretical guarantees but require significantly higher computation compared to simple regularization~\cite{Arad1997OnFT}. Explicitly incorporating additional redundancy by adding nodes and synapses to share the computational load achieves tolerance to single node faults~\cite{248456,315624}.
Additional redundancy like N-Modular Redundancy and resource replication provides partial FT with graceful degradation~\cite{363479}.
Several algorithms modify the training algorithm to generalize the model by adding noise to weights or injecting faults during training which acts as a regularizer~\cite{616152,315624}. However, this regularization is equivalent to mean square error plus Tikhonov function~\cite{317730}. In fact, a theoretical analysis of most of the proposed fault tolerant training algorithms using regularization are equivalent to either Tikhonov and Lasso functions~\cite{5446319}.
Training a NN with Lasso regularization results in sparse parameters as compared to Tikhonov, which allows to prune certain nodes or weights, indicating FT~\cite{8561200}.
The Kullback-Liebler Divergence has also be used as an objective function to improve the regularization of RBF networks~\cite{4439295,8010868}. Alternatively, compression and quantization of models have been shown to increase the fault resilience of NNs~\cite{8782505}. Further, permanent fault detection by thresholding activation values have been considered~\cite{hoang2019ft}.
\section{Fault and Error Model}\label{model}

In this work, we consider the frequently occurring case of stuck at ``0" faults in the nodes (activations) and the parameter (weights and filters) values. There is a particular interest to revisit and scale this fault model for modern NNs due to the emphasis on Neuromorphic hardware and accelerators for the next generation AI making the fault model realistic and relevant. Further, this work considers and evaluates faults in Convolution NNs and thus extending the taxonomy to filter faults. These faults are assumed to occur \textit{after the deployment} of NNs which degrade the model performance over time. Previous research considers the effects of single faults on simple NN topologies with trivial learning tasks which is ineffective for the current state of the art NNs with large number of parameters~\cite{PIURI200118}. To this extent, this work considers \textit{multiple and concurrent faults} randomly occurring in the NN.
Due to large number of parameters and nodes in Deep NNs, exhaustive testing of all possible single faults is prohibitive.
For tractability, this work adopts the strategy of randomly testing for a fraction of stuck at ``0" faults and simulating the effect by measuring the corresponding error in NN prediction. This yields FT estimates that are statistically very close to those obtained by exhaustive testing~\cite{363479}. 
In this work, the focus is on designing Fault Tolerant NNs for the generic node and parameter stuck at ``0" faults which is the same setting followed by other works with which we compare our performance~\cite{8561200,7862272,10.1162/neco.1995.7.1.108,PIURI200118}. To simulate the stuck at ``0" faults in nodes and parameters in the network, binary masks are multiplied with the weights and activations. First, the percentage of total nodes/parameters to remove for each layer is taken as an input to generate the mask which is multiplied with the corresponding weights (parameter faults) or activations (node faults) in the network.

\noindent\textbf{Source of Faults.}  To optimize dataflow in hardware during dot-product between the weights of the current layer with the output from the previous layer, NN accelerators include additional memory along with each processing unit to reuse weights~\cite{8114708}. Inability to access these memory units due to hardware failures results in reading a ``0" value for the corresponding node activations and parameters which occur as errors in the parameters of the NNs~\cite{Li:2017:UEP:3126908.3126964,Temam:2012:DAE:2337159.2337200}.
On the other hand, \textit{node faults} manifest as errors in the activations values in the intermediate layers due to a fault in the processing unit itself resulting in ``0" output. The fault model is applicable to both NN accelerators and general purpose CPUs/GPUs which have an array of ALU with register memory. All nodes in a particular layer are mapped to the array of ALUs for parallel execution. A fault in one ALU/memory results in multiple and concurrent stuck at ``0" errors in activations/parameters across several layers for the specific nodes mapped to the faulty ALU. Such multiple concurrent errors in NNs from small number of faults causes performance degradation make the injection rate in Section 6 realistic.

\noindent\textbf{Other Settings.} Identifying hardware critical bit faults and FT in an adversarial setting~\cite{duddu2020fault} are not within our scope. This work focuses on improving the inherent FT of NNs to randomly occurring stuck at ``0" errors in the nodes and the weights which is a more applicable for IoT applications~\cite{8013784}.
\section{Experiment Setup}\label{reg}

\noindent\textbf{\underline{Motivation: Why does regularization enhance FT?}} FT is exhibited by models with high generalization and regularization functions play an important role in enhancing FT~\cite{7862272}.
To achieve highly generalized model, regularization functions penalise large parameter values and ensure that the models do not overfit.
While a theoretically the relation between generalization and FT has been indicated~\cite{8561200,7862272}, the empirical study of regularization for FT is not fully understood. To this extent, we evaluate the impact of regularization on FT which motivates our proposed approach to enhance the inherent FT of NNs.

\begin{figure}[!htb]
\includegraphics[width=6.5cm]{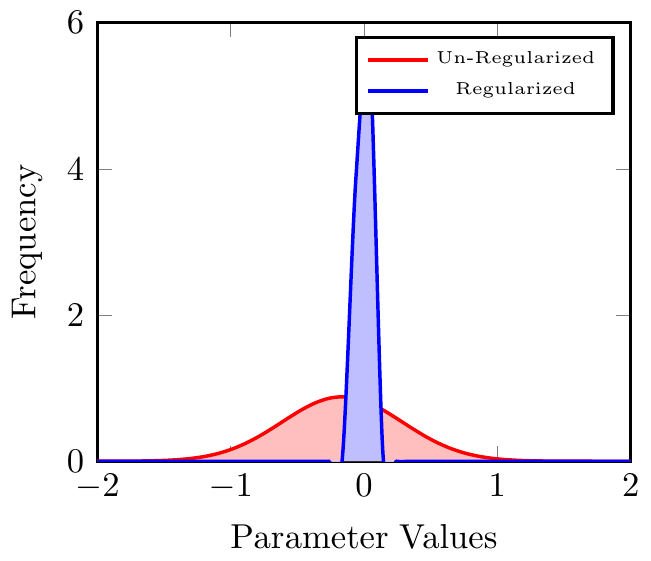}
\caption{\textbf{Effect of Regularization on Parameter Distribution.} The distribution of parameters for unregularized model has a higher standard deviation and on regularization, the standard deviation of the distribution decreases.}
\label{fig:weightdist}
\end{figure}

\begin{figure*}[!htb]
\minipage{0.32\textwidth}
  \includegraphics[width=\linewidth]{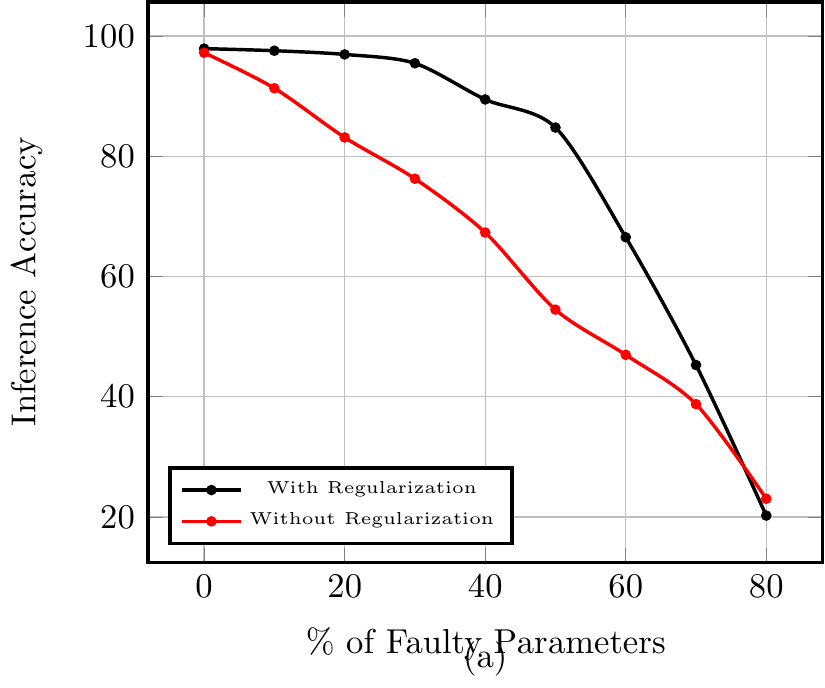}
\endminipage\hfill
\minipage{0.32\textwidth}
  \includegraphics[width=\linewidth]{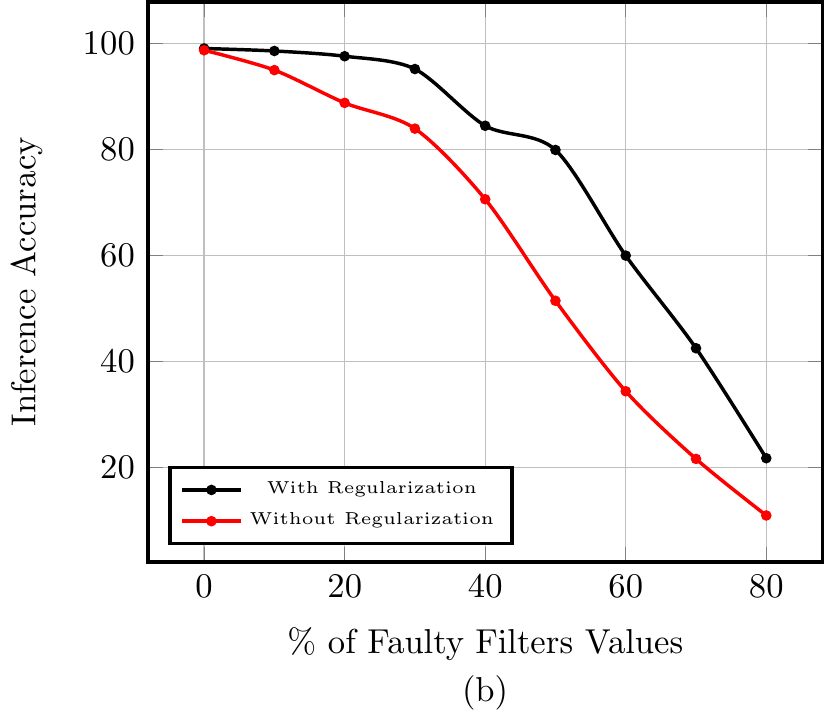}
\endminipage\hfill
\minipage{0.32\textwidth}%
  \includegraphics[width=\linewidth]{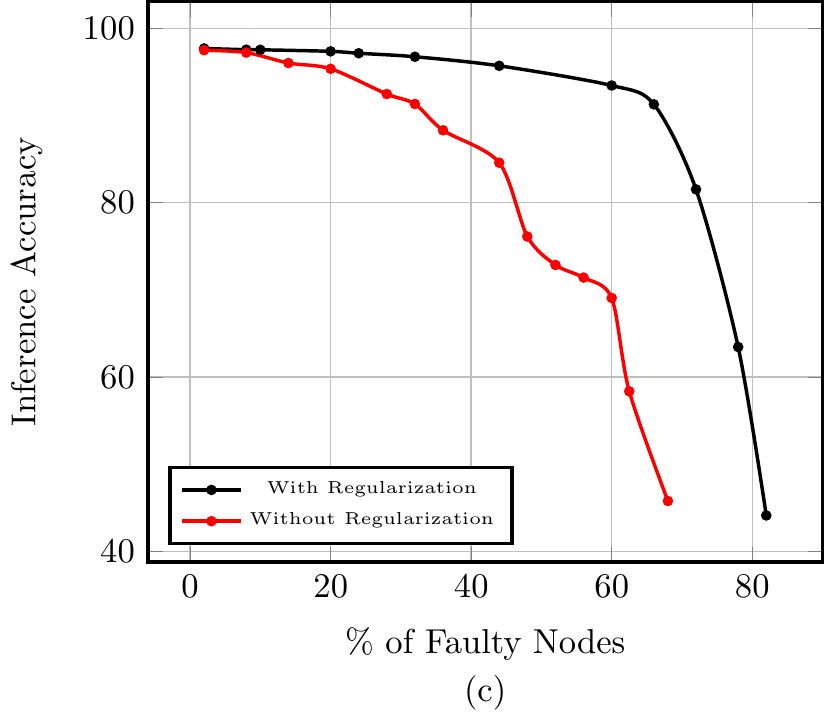}
\endminipage
\caption{\textbf{Comparing the Tolerance of Regularised and Un-Regularised (overfitting) Models.} Performance Degradation of Model Accuracy in presence of (a) Weight Faults in Multilayer Perceptron (b) Filter Faults in Convolution Neural Networks and (c) Node Faults.}
\label{fig:arch_plain}
\end{figure*}

\noindent\textit{\underline{Experiment.}} In order to understand the effect of regularization on FT empirically, we plot the parameter distribution for regularized and overfitting model (Figure~\ref{fig:weightdist}). The two NNs are trained with same architecture: one with regularization (generalized) while the other without regularization (overfitting).

\noindent\textit{\underline{Observation.}} The information in the NN in the form of parameter values follow a Gaussian distribution (Figure~\ref{fig:weightdist}). The parameters after gradient descent optimisation with regularization have similar values, i.e, the parameter distribution has low variation with standard deviation of 0.04772. On the other hand, the parameters of overfitting model after training are significantly more varied which is indicated by the distribution having a larger standard deviation of 0.452404.

\noindent\textit{\underline{Hypothesis.}} Regularization uniformly distributes the weights across all the parameters which ensures that all the nodes are given equal importance during the computation. In the presence of faults, other nodes or synapses can take over the computation without loss of accuracy. In case of overfitting, the model parameters adapt themselves specifically to the training data where some of the nodes or synapses are given more preference compared to others.
Hence, loss of these important nodes or parameters results in a significant drop in performance.

\noindent\textit{\underline{Validation.}} In order to understand the impact more concretely, we simulate stuck at ``0" faults for both the models for three cases: (a) weight faults for Fully Connected Layers, (b) filter faults for convolutional layers and (c) node faults (Figure~\ref{fig:arch_plain}).
The performance degrades gracefully for models with regularization as compared to models without regularization.
In case of fault in parameters, a loss of 50\% of the total nodes result in an accuracy of 54.49\% for overfitting model compared to 84.78\% in regularized models.
In case of convolutional filter faults, a loss of 50\% of the filter result in an accuracy of 51.43\% for overfitting models compared to 79.90\% in regularized models.
For node faults, a 93.46\% accuracy was observed for regularized models as compared to 69.08\% accuracy of overfitting models when 60\% of the nodes are not functioning.
Despite a significant number of faults injected into the system, the performance of the generalized model is still high and operable, compared to models without regularization.

\noindent\underline{\textbf{Metrics: Quantifying FT.}} Since, generalization and FT are strongly correlated, instead of measuring the FT using Section 2.2 which is empirically hard to estimate, we propose a practical estimate for FT using \textit{generalization error}. Given a population of data and corresponding labels, $P(X,Y)$, the goal of training a machine learning model is to minimize the expected loss computed over the entire population $E_{pop} = \mathbb{E}_{(x,y) \sim P(X,Y)} [l(f(x),y)]$. However, this error cannot be exactly computed as it requires knowing the entire distribution of data $P(X,Y)$.
Instead, the loss is computed and minimized over training dataset $D \subset P(X,Y)$ given by $E_{train} = \frac{1}{|D|} \sum_{(x,y) \sim D}(l(f(x),y))$. The difference between the empirical loss over the entire population and the expected loss over the training data helps to identify the extent of model's generalization.
To compute the performance on unseen data, the population error($E_{pop}$) is approximated by computing the expected error over an unseen test data $E_{test} = \frac{1}{|D_{test}|} \sum_{(x,y) \sim D_{test}}(l(f(x),y))$. For a large number of samples $n$, $\lim_{n \to \infty} E_{test} = E_{pop}$.
To compute the degree of overfitting, the difference between the training accuracy ($R_{train}$) and the testing accuracy ($R_{test}$) is measured corresponding to the respective training and testing error: $G_{error} = R_{train} - R_{test}$.
A higher generalization error percent $G_{error}$ indicates more overfitting and the model is prone to significant performance degradation in presence of random faults. \textbf{In summary, we use generalization error to measure FT and test accuracy to estimate the performance degradation in presence of faults.}

\noindent\underline{\textbf{Baselines: State of the Art Regularization.}} Current Deep NNs rely heavily on variants of Lasso or Tikhonov regularization which have been extensively studied to enhance FT~\cite{317730,5446319,8561200}.
While these regularization functions generalize the model by clipping large parameter values, there exists a tradeoff between the test accuracy and the degree of generalization and hence, the resultant models are not maximally tolerant to faults.
As seen Table~\ref{tab:l2reg}, despite having a model with high accuracy, the resulting generalization error for the NNs is high (overfitting), while, for strongly regularized models the test accuracy degrades.
Finding the optimal values of the hyperparameters which results in good accuracy with lower generalization error is a challenging search problem over the large hyperparameter space. Given the input data samples from the training dataset $(x_i,y_i)$ $\in$ $D$ , the objective function to optimize for model $f()$ is computed as $L = \sum_i (y_i - f(x_i)) + \lambda R(\theta_j)$, i.e, the error in prediction of the model from the true label $y$ for the input data point $x$ and regularization function $R()$ to penalise large values of parameters $\theta$.
In case of Lasso regularization, the resulting regularization function is the absolute sum of the individual parameters $\theta$, $L = \sum_i (y_i - f(x_i)) + \lambda \sum_j |\theta_j|$ where $\lambda$ is the regularization hyperparameter to control the scaling for penalising large parameters.
In case of Tikhonov regularization, the loss includes the sum of the square of the parameter values, $L = \sum_i (y_i - f(x_i)) + \lambda \sum_j |\theta_j|^2$.

\begin{table}[!htb]
\caption{\textbf{Generalization Accuracy Trade-off.} Regularisation results in a tradeoff between test accuracy and generalization error.}
\centering
\begin{tabular}{| l | c c c |}
\hline
	\textbf{Regularisation}  & \textbf{Training}  & \textbf{Testing}  & \textbf{Generalization} \\
	\textbf{Hyperparameter} &  \textbf{Accuracy}  & \textbf{Accuracy} & \textbf{Error}\\
	\hline
	\multicolumn{4} {|c|} {Tikhonov Regularization}\\
	\hline
\textbf{0} &  $98.60\%$ & $88.90\%$ & $9.70\%$  \\
\textbf{0.0001} &  $97.84\%$ & $89.89\%$ & $7.95\%$ \\
\textbf{0.001} &  $93.88\%$ & $88.86\%$ & $5.02\%$ \\
\textbf{0.1} &  $66.57\%$ & $65.74\%$ & $0.83\%$  \\
	\hline
\multicolumn{4} {|c|} {Lasso Regularization}\\
\hline
\textbf{0} &  $96.77\%$ & $89.53\%$ & $7.24\%$  \\
\textbf{0.0001} &  $90.26\%$ & $86.70\%$ & $3.56\%$  \\
\textbf{0.001} &  $85.07\%$ & $84.20\%$ & $0.87\%$  \\
\textbf{0.01} &  $70.49\%$ & $70.37\%$ & $0.12\%$  \\
\hline
\end{tabular}
\label{tab:l2reg}
\end{table}

Tikhonov regularization penalises large parameters more strongly as compared to Lasso, while Lasso enhances FT by incorporating sparsity~\cite{8561200}. Further, the theoretical analysis of most of the proposed fault tolerant training algorithms using regularization are equivalent to either Tikhonov and Lasso functions~\cite{5446319} and these techniques depict a trade-off between generalization and accuracy~\cite{7862272}. This is also applicable to dropout which has extensively been shown to approximate scaled Tikhonov regularizer or noise added to parameters~\cite{10.5555/2999611.2999651,NIPS2017_7096,DBLP:journals/corr/abs-1905-11320,6797856}.
Tikhonov and Lasso functions have been extensively studied for enhancing the FT of NNs for a wide range of applications along with theoretical guarantees of convergence. \textbf{In this work, we use Tikhonov and Lasso functions as our baseline and compare them with our novel fault tolerant training algorithm based on game theoretic guarantees.}

\section{Adversarial Fault Tolerant Training}\label{advfault}

The objective of the proposed algorithm is to enhance the inherent FT to random stuck at ``0" faults by strongly regularizing the model while ensuring high accuracy and providing theoretical guarantees.
The NN, separated into a FE and a FCC Network, is trained in two phases with different objective functions to achieve the maximal performance \textit{and} generalisation.
The FE is trained in Phase I with an unsupervised training which is further modelled as two games. In Phase II, the FE is attached to the FCC and further fine-tuned with supervised learning objective to perform classification tasks.

\subsection{Phase I: Unsupervised Training}

\begin{figure*}[!htb]
\includegraphics[width=0.7\textwidth]{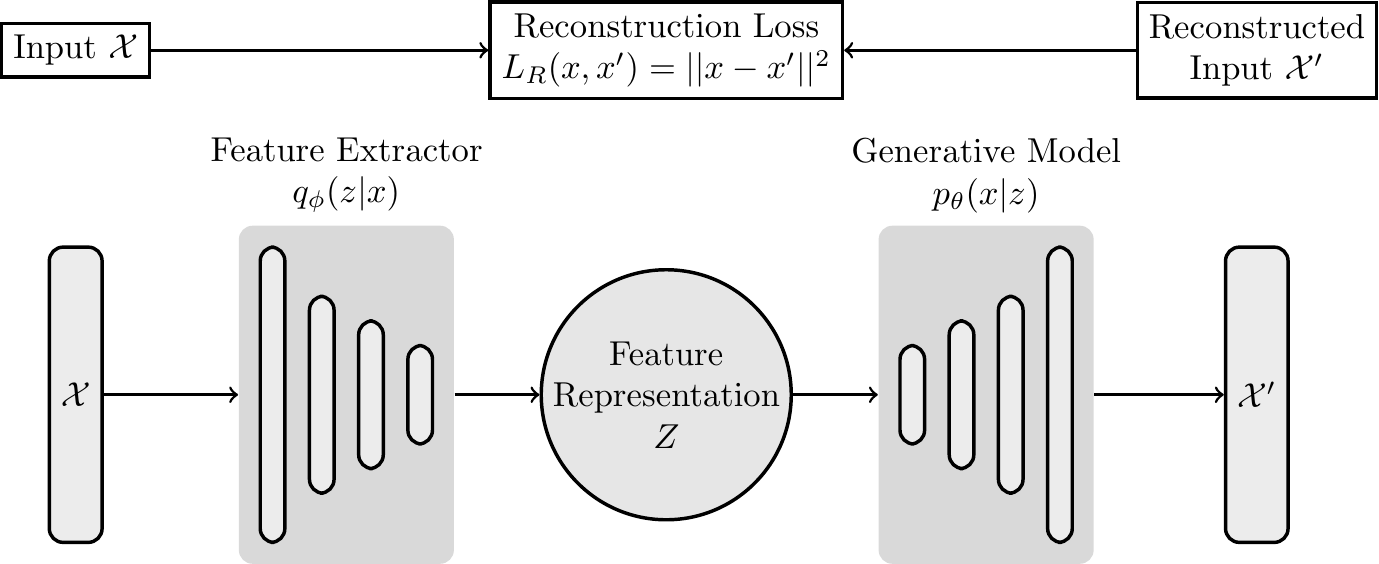}
\caption{\textbf{Phase I: Game 1. Unsupervised Training for Robust Features} The FE is trained with a Generative Model $p_{\theta}$ to extract robust intermediate feature representation of the images by minimising the reconstruction loss. }
\label{fig:attack}
\end{figure*}

The FE has an unsupervised learning objective of mapping the input images to feature space representation. This requires FE to \textit{extract robust features} given the image and ensure \textit{distributional smoothness} of the extracted features close to a prior Gaussian distribution. Distribution smoothness is essential to ensure that mapping the feature representation back to image is distributionally close to the original image~\cite{makhzani2015adversarial}.
To achieve this, the Extractor is trained by solving two adversarial games in the presence of NNs with conflicting objectives where Game 1 trains FE for robust features while Game 2 achieves the objective of distributional smoothness.

\noindent\ding{202} \textbf{\underline{Game 1: Feature Extractor vs. Generative Network.}} The objective of this game is to train the FE to identify and extract the dominant features in an image.
The game is played between the FE ($q_{\phi}()$) and a Generative Network ($p_{\theta}()$) with conflicting objectives.
The goal of the FE is identify and extract the dominant features $q_{\phi}(x)$ given an input image ($x$).
On the other hand, Generative reconstructs the input image ($x'$) given the features $q_{\phi}(x)$ extracted from the actual image $x$, given by $p_{\theta}(q_{\phi}(x))$.
The two models are trained simultaneously to minimize the loss computed in the reconstruction process and learn the identity function $p_{\theta}(q_{\phi}(x))$ = $x$ using the FE as image encoder and Generator as decoder. The models train and update their parameters to choose the best possible strategy where the reconstructed image from the generated features is close to the original input image~\cite{makhzani2015adversarial}.
Ideally, after multiple epochs of training and parameter updates of both the adversary networks, the reconstructed image is indistinguishable for the original input image ensuring that FE extracts dominant and robust image features which can be efficiently reconstructed by the Generator.

\noindent\textbf{Theoretical Analysis.} The reconstruction loss between $x$ and $x'$, $L_R (x, x')= ||x - x'||^2$, is minimized using Stochastic Gradient Descent to tune the parameters in the direction of minimum reconstruction loss. The gradients are computed first to update the Generative Model parameters $\theta$ as $\nabla_{\theta} ||x - x'||^2$ updated as $\theta = \theta - \alpha \nabla_{\theta} ||x - x'||^2$ with learning rate of $\alpha$. Then, the parameters of the FE are then updated similarly by computing the gradients of loss function with respect to the $\phi$, $\nabla_{\phi} ||x - x'||^2$. The parameters are iteratively updated till the minima of the loss function is achieved. The convergence to a minima using reconstruction loss follows that of Gradient Descent algorithms~\cite{arora2018convergence}.

\begin{figure*}[!htb]
\includegraphics[width=0.7\textwidth]{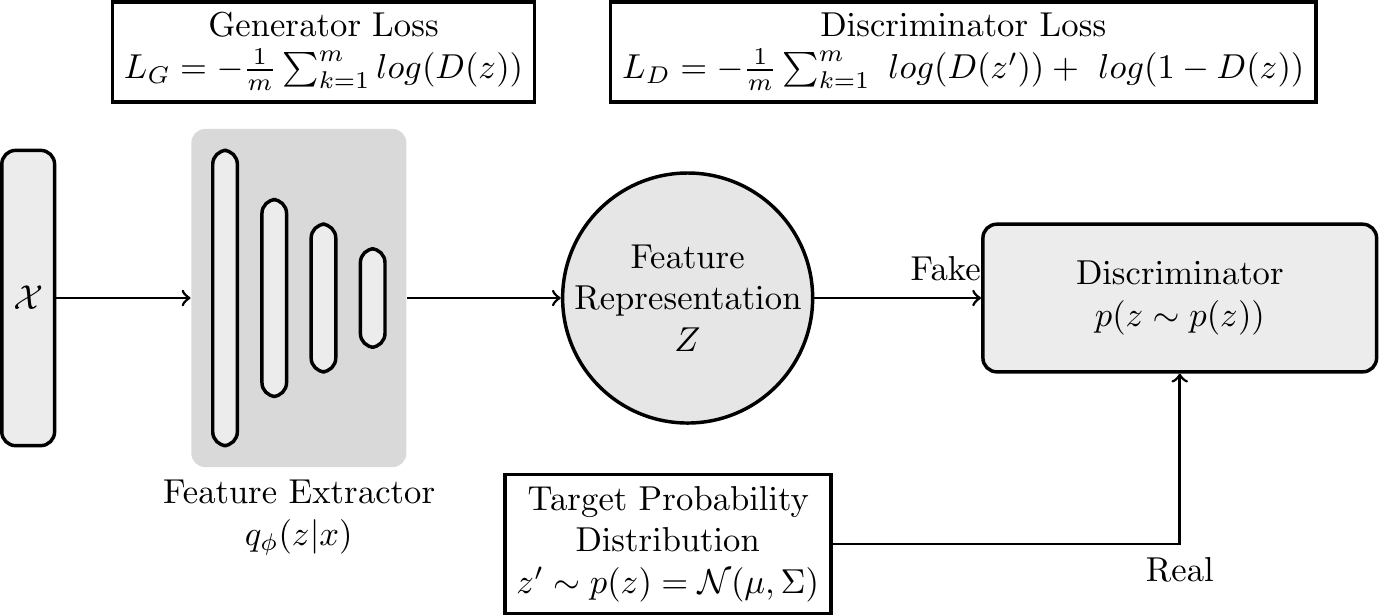}
\caption{\textbf{Phase 1: Game 2. Unsupervised Pre-Training for Regularising Feature Extractor.} The Feature Extractor is adversarially trained by solving a min-max game with the Discriminator to ensure the distribution of feature space matches the target prior distribution.}
\label{fig:attack}
\end{figure*}

\noindent\ding{203} \textbf{\underline{Game 2: Feature Extractor vs. Discriminator.}} The objective of the second game is distributional smootheness of the feature space by training the FE to output image features following a prior Gaussian distribution to ensure there are no gaps in the feature space~\cite{makhzani2015adversarial}.
This optimisation is solved as a zero-sum non-cooperative (minimax) game between two NNs where both the networks choose the optimal strategy to reduce the success of the other network's objective.
Since, both the Networks are playing the game against each other, a Nash Equilibrium exists where both players take their best decision decision~\cite{NIPS2014_5423}.
The goal of the Discriminator Network ($D()$) is to distinguish between ``fake" data points ($z$) generated from the FE and ``real" data points ($z'$) sampled from prior distribution.
The FE \textit{minimizes} the difference between generated features and the target prior distribution such that the Discriminator Network error is \textit{maximised}.
The Discriminator receives $z$ sampled from FE's output $q_{\phi}(z|x)$ and $z'$ sampled from the true prior $p(z)$ and assigns a probability to each sample of coming from $p(z)$.
The gain of the discriminator due to correctly distinguishing between the data points sampled from different distribution is computed as

\begin{equation}\label{discloss}
G_D(z,z') = - \frac{1}{m} \sum _{k=1} ^m \ log(D(z')) + \ log(1 - D(z))
\end{equation}
where $m$ is the minibatch size, $z$ is generated by the encoder and $z'$ is a sample from the true prior.
For the FE, the goal is to maximise the errors (or minimize the gain) of the Discriminator by generating samples similar to the prior distribution samples while the Discriminator tries to maximise the overall gain.
In other words, the FE trains to minimize the maximum gain of the Discriminator Network.
The loss for the FE is computed as,

\begin{equation}\label{genloss}
L_G(z) = - \frac{1}{m} \sum _{k=1} ^m log(D(z))
\end{equation}

The loss computed is back-propagated through the discriminator $D()$ to update its weights followed by which the FE updates its parameters.
The FE eventually generates samples $z \sim q_{\phi}(z|x)$ close to the target prior $p(z)$ such that the discriminator cannot distinguish between the two inputs (random guess strategy).
Over time, the loss of the discriminator increases while the FE minimizes the maximum gain of the Discriminator. To solve the min-max optimisation, in each epoch of training, the FE and Discriminator are alternatively trained to select the best strategy against the other player (Algorithm~\ref{alg:training}).

The overall \textit{adversarial loss} of the FE used to update the parameters includes both the reconstruction loss and the loss given by the discriminator network ($D$) as shown below,
\begin{equation}\label{minimax}
\min_{q_{\phi}} \left(L_R(x,x') + \max_{D} G_D(z,z') \right)
\end{equation}
where the model is updated with the reconstruction loss first, followed by the update over the discriminator loss. Hence, Game 1 and Game 2 are solved alternatively in one iteration of training.

\noindent\textbf{Theoretical Analysis.} The Discriminator is first is trained to maximise the distinction between the data points sampled from feature space distribution (attributed as ``fake") and sampled from a chosen prior distribution (attributed as ``real") (inner maximisation in Equation ~\ref{minimax}).
The FE on the other hand is trained to minimize the maximum deviation given in Equation~\ref{genloss}.
Both the networks are jointly trained to attain the equilibrium (saddle) point using Stochastic Gradient Descent algorithm.

\begin{figure*}[!htb]
\minipage{0.32\textwidth}
  \includegraphics[width=\linewidth]{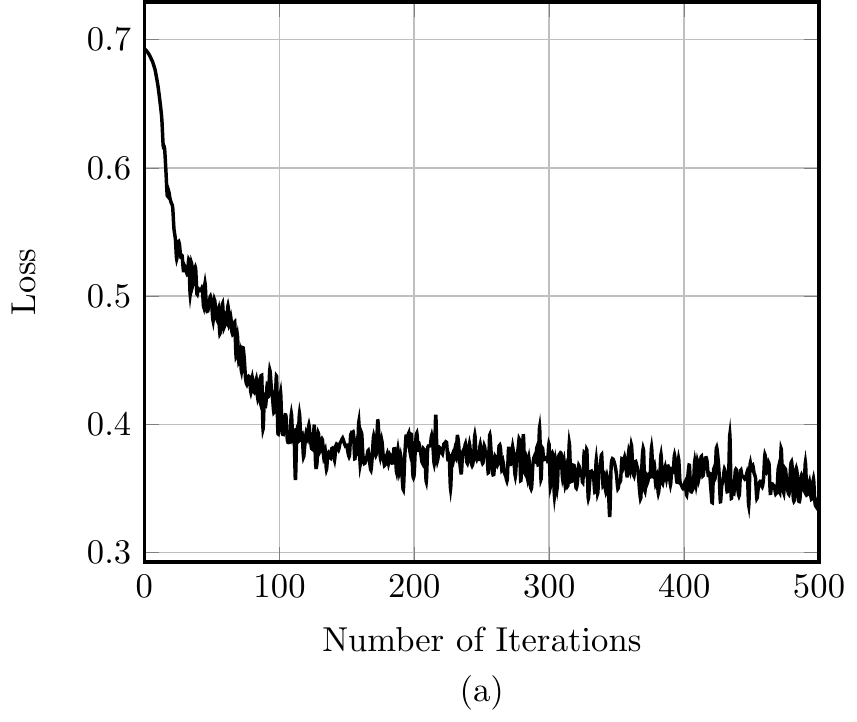}
\endminipage\hfill
\minipage{0.32\textwidth}
  \includegraphics[width=\linewidth]{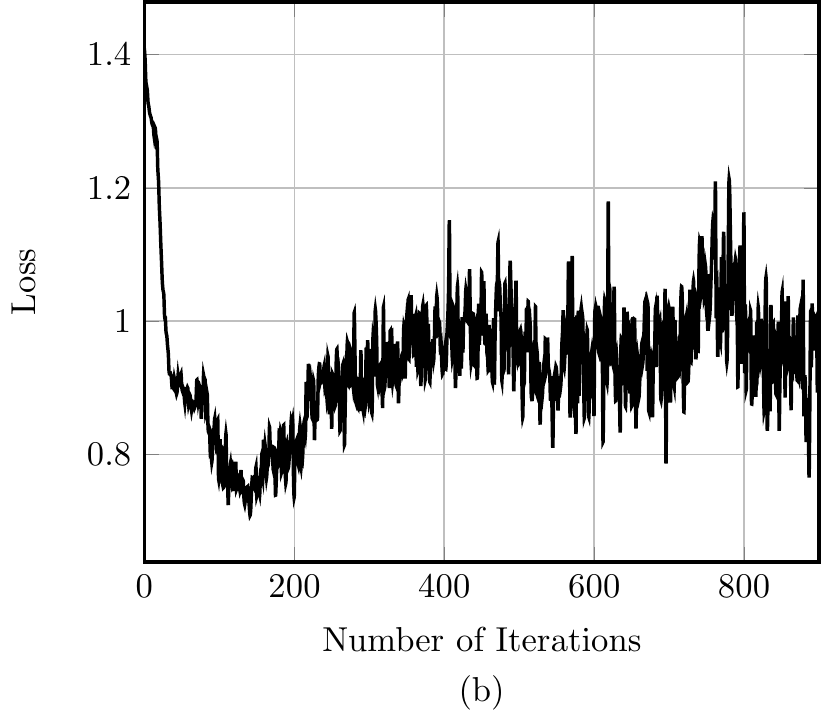}
\endminipage\hfill
\minipage{0.32\textwidth}%
  \includegraphics[width=\linewidth]{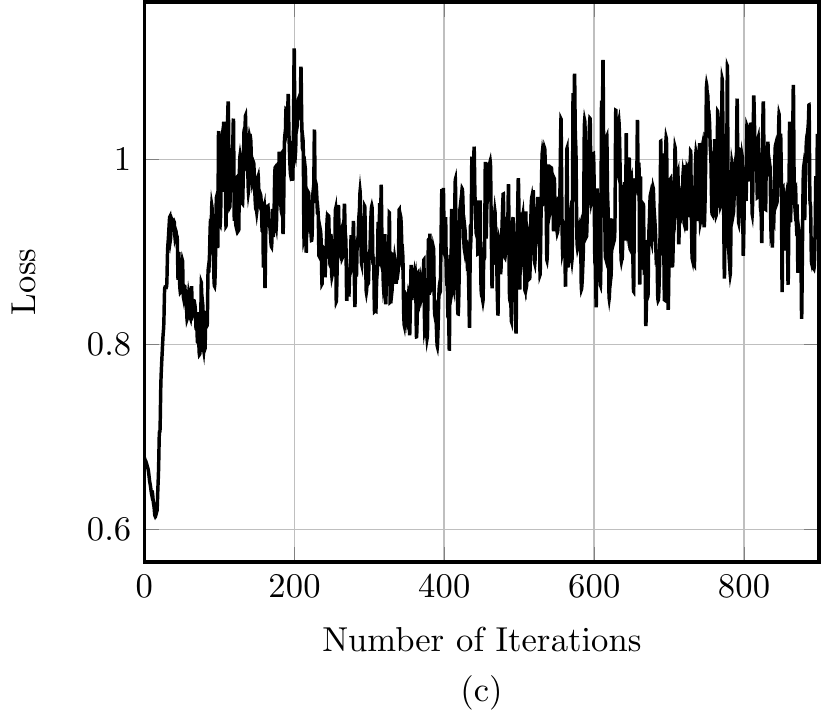}
\endminipage
\caption{\textbf{Loss Functions.} (a) Minimizing the Reconstruction loss; (b) Minimising FE loss and (c) Maximising the gain of Discriminator during the min-max optimisation.}
\label{fig:loss}
\end{figure*}

\begin{multline}
\min_{q_{\phi}} \max_{D} V(q_{\phi}, D) \\
= \mathbb{E}_{x \sim p_{z}}[log(D(x))] + \mathbb{E}_{x \sim q_{\phi}(z | x)}[1-log(D(G(x)))] \\
 = \int_{x} p_{z}(x)log(D(x))dx + p_{q_{\phi}}(x)log(1-D(x))  dx\\
\end{multline}

On differentiating the function inside the integral and equating to zero, the following optimal value  of the Discriminator is obtained given that the FE ($q_{\phi}()$) is fixed,

\begin{equation}
D^*_G(x) = \frac{p(z)}{p(z) + q_{\phi}(z | x)}
\end{equation}

As an extension of Theorem 1 in \cite{NIPS2014_5423}, it can be shown that the min-max game has a global optimum for $p(z) = q_{\phi}(z | x)$, i.e, the prior distribution is same as the feature space distribution.
Since, the global optima is achieved at $p(z) = q_{\phi}(z | x)$, at the point of global optimum, the Discriminator $D^*(x)=\frac{1}{2}$, i.e, the model cannot distinguish (random guess) between the data points sampled from the true prior distribution and feature space distribution~\cite{NIPS2014_5423,makhzani2015adversarial}.
The proof of convergence to equilibrium of min-max game (Algorithm~\ref{alg:training}) follows directly from Proposition 2 in~\cite{NIPS2014_5423}.

\noindent\textbf{FT as Min-Max Constraint Optimisation.} Minimax constraint during NN helps to generalise the model by minimizing the maximum deviation between the error computed over fault-less Network $\mathcal{H}_\mathcal{N}(\mathcal{X}$) and Faulty Network $\mathcal{H}_{\mathcal{N}_{fault}}(\mathcal{X}$)~\cite{105414,712162}.
The minimax constraint trains an $\epsilon$-Fault Tolerant Network to minimize the threshold $\epsilon$ for maximum possible deviation, $\mid \mathcal{H}_\mathcal{N}(\mathcal{X}) -  \mathcal{H}_{\mathcal{N}_{fault}}(\mathcal{X}) \mid$ $<$ $\epsilon$.
In other words, the goal is to determine the minimum value of $\epsilon$ such that the resultant weights of $\mathcal{N}$ maintains a high accuracy and additionally satisfies $\epsilon$-FT.
Since, the minimax constraint optimisation is highly non-linear and non-differentiable, previous approaches approximate it~\cite{712162} or solve it using quadratic programming results in partial FT and computationally more expensive and not scalable to current architectures~\cite{105414}.
In this work, the minimax optimisation is efficiently solved and modelled as a game between two adversary NNs with conflicting objectives. The Discriminator Network maximises the deviation between the feature space distribution (from faulty network) and the prior target distribution (from fault-less network) while FE minimizes the maximum deviation computed by the Discriminator.
This game theoretic solution to minimax solution is scalable to large NNs and incorporates FT as a constraint during training in the features pace instead of output space as seen in prior work.

\subsection{Supervised Training}

\begin{figure}[!htb]
\includegraphics[width=6.5cm]{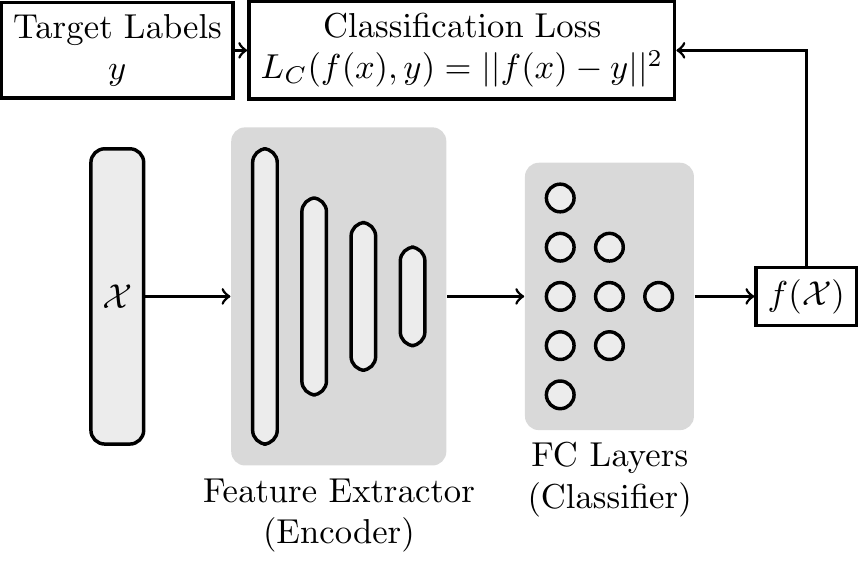}
\caption{\textbf{Phase II. Supervised tuning of FE + FCC Model to minimize the classification loss.}}
\label{fig:advclassifier}
\end{figure}

\begin{algorithm*}[htb!]
	\caption{\textbf{Adversarial Fault Tolerant Training Algorithm.} In Phase I, the unsupervised training of FE $\phi$ updates the parameters by minimizing the reconstruction loss (Game 1). The FE is simultaneously trained with Discriminator (Game 2) to minimize the maximum deviation between the feature space and Gaussian distribution. In Phase II, the FE is attached to the FCC for fine-tuning for supervised task by minimizing the classification loss.}
	\begin{algorithmic}[1]
	\STATE \textbf{\underline{Phase I}}
		\FOR{unsupervised training epochs } 
		\vspace{0.5cm}
		\STATE \ding{202} \textbf{\underline{Game 1}}
		\STATE \textbf{(1) Train the Feature Extractor to extract robust features with Generator by minimizing reconstruction loss.}
		\vspace{0.2cm}
		\STATE Sample $\{x^{(i)}\}_{i=1}^m \sim \mathcal{P}(X,Y)$
		\STATE Compute $z^{(i)} = \text{q}_{\phi}(x^{(i)})$
		\STATE Compute $x'^{(i)} = \text{p}_{\theta}(x^{(i)} | z^{(i)})$
		\STATE Update the Feature Extractor and Generator using \textit{Stochastic Gradient Descent} over the parameters $\theta$ and $\phi$
		\STATE Backpropagation  loss, $\mathcal{L}_{\text{rec}} = || x - x'||^{2}$

		\vspace{0.5cm}
        \STATE \ding{203} \textbf{\underline{Game 2}}
		\STATE \textbf{(2) Train Discriminator to distinguish Gaussian and generated samples (Keep Feature Extractor Fixed).}
		\vspace{0.2cm}
		\FOR{k steps}

		\STATE Sample data points from true prior distribution $z'^(i) \sim p(z)=\mathcal{N}(\mu,\Sigma)$
		\STATE Compute $z^{(i)} = \text{q}_{\phi}(x^{(i)})$
		\STATE Update the Discriminator using \textit{Stochastic Gradient Ascend}
		\STATE Backpropagation Loss: $\mathcal{L}_D(z,z') = - \frac{1}{m} \sum _{k=1} ^m \ log(D(z')) + \ log(1 - D(z))$

		\ENDFOR
		\vspace{0.5cm}
	  \STATE \textbf{(3) Train the Feature Extractor for distributional smoothness of features (Keep Discriminator fixed).}
		\vspace{0.2cm}
		\STATE Update the Feature Extractor $q_{\phi}$ using \textit{Stochastic Gradient Descent} over the parameters $\phi$
		\STATE Backpropagation Loss: $- \frac{1}{m} \sum _{k=1} ^m log(D(z))$

		\ENDFOR

		\vspace{0.5cm}
		\STATE \textbf{\underline{Phase II}}
		\STATE \textbf{(4) Attach Fully Connected Classifier to Feature Extractor followed by supervised fine-tuning.}
		\vspace{0.2cm}
		
		\FOR{supervised training epochs } 

		\STATE Sample data points with corresponding labels  $\{(x^{(i)},y^{(i)})\}_{i=1}^m$ $\sim$ $D$.
		\STATE Update the Classifier Model using \textit{Stochastic Gradient Descent} over its parameters
		\STATE Backpropagation Loss: $\mathcal{L}_C (y, f(x)) = ||f(x) - y||^2$

		\ENDFOR

	\end{algorithmic}
	\label{alg:training}
\end{algorithm*}

The unsupervised training of the FE acts as a strong regulariser~\cite{pmlr-v9-erhan10a}.
The pre-trained FE ($q_{\phi}$), now, is attached to the Classifier Network ($f_{class}$) for supervised training.
Formally, the classifier network maps the input latent space representation of image($z$) to the corresponding class in $Y$, i.e, $f_{class}: q_{\phi}(z|x) \rightarrow Y$ for a given input $x$.
While the parameters of the Classifier Network are trained for predicting the class from the features of an image, the parameters of the FE are updated using the prediction error.
The model parameters are updated by minimizing the \textit{classification loss} computed using the ground truth labels $y$ and the predicted labels $f(x)$ given by, $L_C (y, f(x)) = ||f(x) - y||^2$.

Notably, the unsupervised training ensures that the FE extracts dominant and robust features of the image which is close to a Gaussian distribution. The effectiveness of the unsupervised learning strategy lies in efficiently learning the input distribution $P(X)$ in order to improve the supervised classification $P(Y | X)$ for input space $X$ and corresponding label space $Y$~\cite{pmlr-v9-erhan10a,pmlr-v5-erhan09a}. This results in a strongly regularized network which, as we indicate in the next section, is more effective than standard regularisation schemes like Tikhonov and Lasso.
The detailed algorithm combining the above training strategies is given in Algorithm~\ref{alg:training}.

\section{Evaluation}\label{eval}

In this section, we evaluate our two phase training algorithm on four NNs trained on two standard image classification tasks implemented in Pytorch framework. The code is publicly available for reproducing experiments\footnote{Code: https://gitlab.com/vasishtduddu/FaultTolerantNN}.

\subsection{Datasets}

\noindent\textbf{FashionMNIST.} The dataset consists of a training set of 60,000 examples and a test set of 10,000 examples.
Each data sample is a 28 $\times$ 28 grayscale image from one of 10 classes: shoes, jackets, shirt etc.

\noindent\textbf{CIFAR10.} The dataset contains 60,000 images, with 50,000 images for training and 10,000 images for testing where each data point is 32 $\times$ 32 coloured image. The images are clustered into 10 classes representing different real world objects. This dataset is significantly more complex compared to FashionMNIST.

\begin{table*}[htb!]
\caption{\textbf{Comparison of Accuracy and Generalization for different regularizers.} The proposed training approach results in lower generalisation error while resulting in higher test accuracy. For FashionMNIST, highlighted cell indicates Architecture 1 while the cells below is Architecture 2. For CIFAR10, Architecture 3 is highlighted with Architecture 4 below.}
\centering
\resizebox{2\columnwidth}{!}{%
\begin{tabular}{| l | c c c | c c c |}
	\hline
	 & \multicolumn{3}{c|}{\textbf{FashionMNIST}} & \multicolumn{3}{c|}{\textbf{CIFAR10}} \\
	\hline
	 & \textbf{Training Accuracy} & \textbf{Testing Accuracy}  & \textbf{Generalization Error}  & \textbf{Training Accuracy} & \textbf{Testing Accuracy} & \textbf{Generalization Error}\\
	\hline
	\cellcolor{gray!20}No Regularization &  \cellcolor{gray!20}$99.51\%$ & \cellcolor{gray!20}$89.42\%$ & \cellcolor{gray!20}$10.09\%$ & \cellcolor{gray!20}$96.35\%$ & \cellcolor{gray!20}$86.11\%$ & \cellcolor{gray!20}$10.24\%$ \\
	\cellcolor{gray!20}Lasso &  \cellcolor{gray!20}$98.42\%$ & \cellcolor{gray!20}$88.96\%$ & \cellcolor{gray!20}$9.46\%$ & \cellcolor{gray!20}$85.95\%$ & \cellcolor{gray!20}$77.78\%$ & \cellcolor{gray!20}$8.17\%$ \\
	\cellcolor{gray!20}Tikhonov &  \cellcolor{gray!20}$97.16\%$ & \cellcolor{gray!20}$88.52\%$ & \cellcolor{gray!20}$8.64\%$ & \cellcolor{gray!20}$93.87\%$ & \cellcolor{gray!20}$87.21\%$ & \cellcolor{gray!20}$6.66\%$ \\
	\cellcolor{gray!20}Proposed Approach &  \cellcolor{gray!20}$95.78\%$ & \cellcolor{gray!20}$90.25\%$ & \cellcolor{gray!20}$5.53\%$  & \cellcolor{gray!20}$91.36\%$ & \cellcolor{gray!20}$87.58\%$ & \cellcolor{gray!20}$3.78\%$ \\
	\hline
	No Regularization &  $99.60\%$ & $90.94\%$ & $8.66\%$ & $85.16\%$ & $81.71\%$ & $3.45\%$ \\
	Lasso &  $96.98\%$ & $91.19\%$ & $5.79\%$ & $83.90\%$ & $81.68\%$ & $2.22\%$ \\
	Tikhonov &  $95.74\%$ & $90.61\%$ & $5.13\%$ & $84.83\%$ & $81.71\%$ & $3.12\%$ \\
	Proposed Approach &  $96.00\%$ & $91.55\%$ & $4.55\%$  & $82.96\%$ & $81.77\%$ & $1.19\%$ \\
	\hline
\end{tabular}
}
\label{tab:results}
\end{table*}

\begin{figure*}[!htb]
\minipage{0.32\textwidth}
  \includegraphics[width=\linewidth]{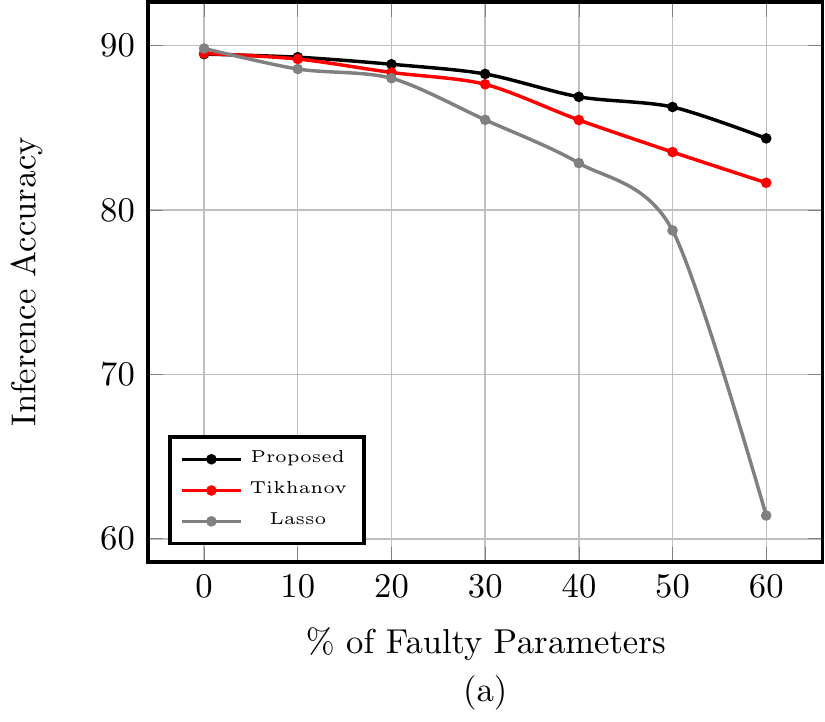}
\endminipage\hfill
\minipage{0.32\textwidth}
  \includegraphics[width=\linewidth]{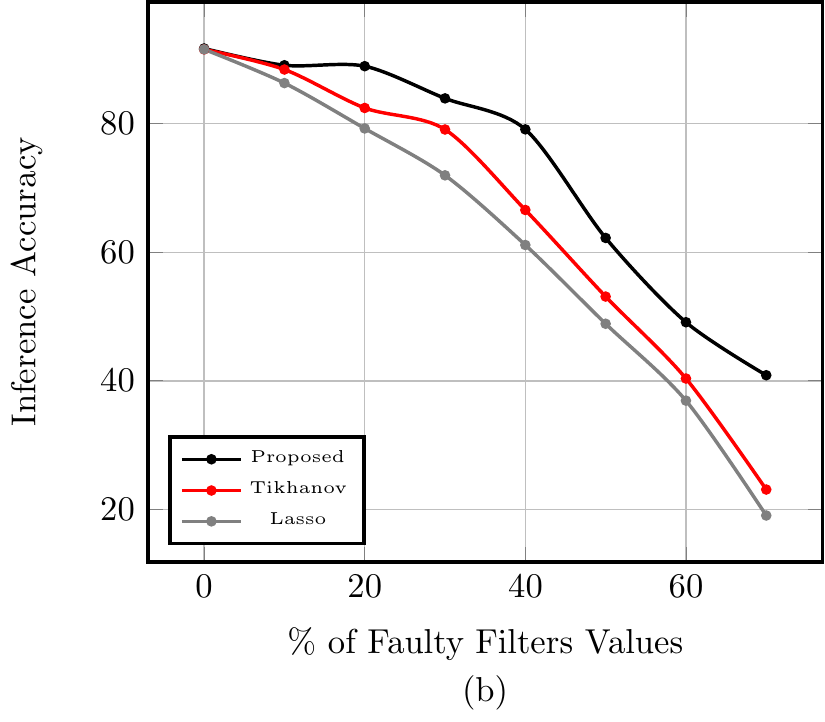}
\endminipage\hfill
\minipage{0.32\textwidth}%
  \includegraphics[width=\linewidth]{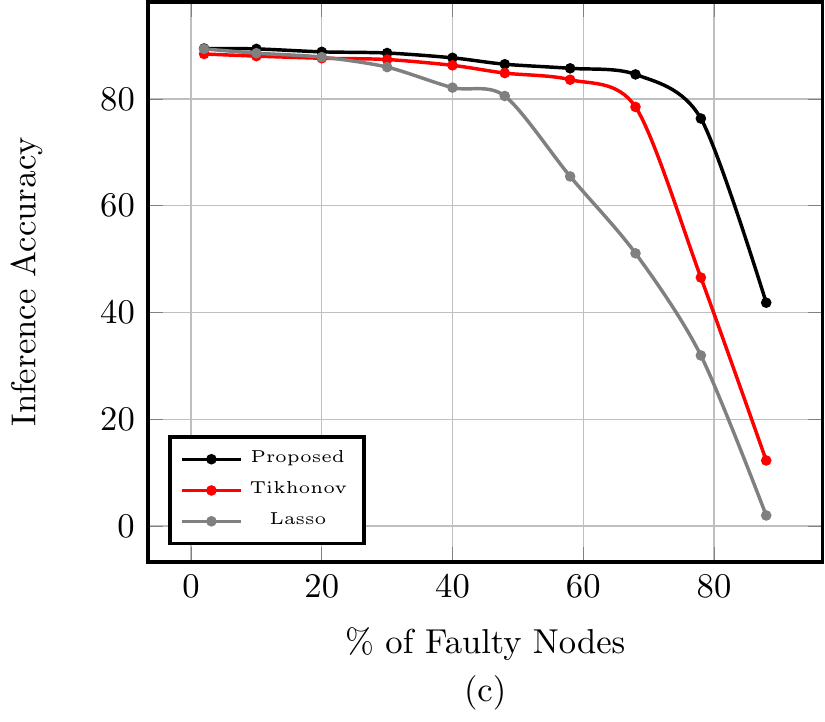}
\endminipage
\caption{\textbf{Comparison of Fault Tolerance for different regularisation.} Performance degradation of the Neural Network models on injecting (a) Weight Faults (b) Filter Faults and (c) Node Faults, trained using proposed technique, Tikhanov and Lasso regularisation. The higher fault tolerance of the proposed approach compared to Tikhanov and Lasso is indicated by the gap in the curves.}
\label{fig:faultdeg}
\end{figure*}

\subsection{Architectures}

For experiments, four architectures with different network depth (number of layers) and complexity are considered to evaluate the scalability of the algorithm. For each dataset, two different architectures are chosen and trained till convergence. Further, the training algorithm is model agnostic and hence, can be used for more complex datasets and architectures.

\noindent\textbf{FashionMNIST.} A Fully Connected Deep NN (\textit{Architecture 1}) and a Convolution NN Architecture \textit{Architecture 2} are chosen for evaluation. The Architecture 1 has three hidden layers of sizes [512, 1024, 512] with ReLu activation while Architecture 2 includes two convolutional layers of 20 and 50 filters with $5\times5$ filters and two maxpool layers of kernel size $2\times$2 and stride of 2. For both models, the FCC comprises of a hidden layer with 512 nodes followed by the output layer of 10 nodes. The Generative Network for both the architectures is a Fully Connected network of size [512, 512, 784].

\noindent\textbf{CIFAR10.} \textit{Architecture 3} includes seven convolutional layers with the number of filters as [64, 128, 256, 512, 256, 128, 128] with LeakyReLu activation with scaling factor of 0.1. \textit{Architecture 4} is a smaller network with three convolutional layers with number of filters as [64, 128, 128], kernel size of 3$\times$3 and stride of 2.
Both the networks use FCC with 1024 nodes and output layer of 10 nodes.
The Generative Network for both the architectures has four convolutional layers, each followed by an upsampling layer with a bilinear scaling factor of 2.

\noindent\textbf{Discriminator Network.} The Discriminator is common to both the datasets and includes a simple multilayer perceptron with two hidden layers of 512 nodes each followed by the output node for binary classification of ``Fake" vs ``Real" data distributions.
The Discriminator architecture is trained using a learning rate of $5e-5$ along with Dropout regularization.
CIFAR10 is adversarially trained with a Discriminator for 10k iterations while FashionMNIST was trained for 2k epochs.

\subsection{Adversarial Framework}

The effectiveness of the proposed training to improve the generalization of NN classifiers is evaluated and addressed in this section.
A comparison of generalization for the NNs using the proposed approach with state of the art regularizers is shown in Table~\ref{tab:results}. The extent of overfitting in the model is measured using the generalization error and is for the best model performance obtained after finding optimal hyperparameters using Grid Search.
For FashionMNIST dataset, unregularized \textit{Architecture 1} has about 1.82x lower generalization as compared to model trained using the proposed algorithm, while \textit{Architecture 2} shows 1.9x higher generalization.
In case of CIFAR10 architectures, \textit{Architecture 3} shows a higher generalization of about 2.7x using the proposed approach while about 3x superior generalization in \textit{Architecture 4} compared to the corresponding unregularized model.
In case of Tikhonov, the proposed approach has a lower generalization error by 3.11\% for Architecture 1, while a 2.88\% lower generalization in case of Architecture 3.
On the other hand, for Lasso, the Adversarial training results in 3.93\% lower generalization error in Architecture 1 and 4.39\% for Architecture 3. Similar results are observed for Architecture 2 and 4
where the proposed approach has lower generalization error and higher accuracy compared to the other regularization functions while ensuring higher accuracy.

\subsection{Performance in the Presence of Faults}

We further validate our approach by simulating random faults to evaluate the performance degradation and resilience of model after deployment (Figure~\ref{fig:faultdeg}).
The metric used to determine the performance is the test accuracy on data samples that the NN has not previously seen.
The evaluation considers weight faults for Fully Connected layers, filter faults for convolution layers and node faults for the Fully Connected Layers. We observe that the NNs trained using the proposed training algorithm results in higher FT compared to models trained using Lasso and Tikhanov. In case of node faults, for a 68\% faults in nodes in the all hidden layers, the model performance is 84.60\% for proposed algorithm, compared to 78.51\% (Tikhanov) and 51.09\% (Lasso).
Further, for the proposed approach, the performance drop is just 4.8\% compared to 9.92\% (Tikhanov) and 38.29\% (Lasso).
In case of weight faults, the accuracy is 84.36\% for proposed approach compared to 81.66\% (Tikhanov) and 61.42\% (Lasso) for a 60\% parameter faults.
For filter faults, the NNs can tolerate upto 40\% filter faults with 12.58\% accuracy drop (proposed) compared to 24.98\% (Tikhnov) and 30.44\% accuracy (Lasso).
This indicates the \textit{superior} FT of the NNs trained using the proposed approach.

\section{Conclusions}\label{conclusions}

Enhancing the inherent FT of NNs through regularisation using simple functions is insufficient due to the trade-off between the model classification accuracy and generalisation.
In this work, a novel training algorithm combining both unsupervised and supervised learning to improve generalisation and attain superior FT is proposed, using two different components of the Neural Networks, Feature Extractor and Classifier Network, based on the difference in functionality and learning objective.
Feature Extractor is trained using unsupervised learning paradigm which is modelled as two simultaneous games in the presence of adversary networks with conflicting objectives to the Extractor.
This strategic training strongly regularises the Feature Extractor which is attached to the Classifier Network for supervised tasks.
The resultant Neural Networks display superior FT as compared to commonly adopted state of the art regularisers, Tikhonov and Lasso.
The proposed FT training Algorithm is model agnostic and can scale to larger datasets and complex architectures. The proposed algorithm can be readily used for practical applications requiring deployment of Deep NNs.

\section*{Acknowledgements}
Research supported by the European Research Development Fund under the Competitiveness Operational Program (BioCell-NanoART = Novel Bio-inspired Cellular Nano-architectures, POC-A1-A1.1.4-E nr. 30/2016)


\end{document}